\documentclass[journal]{IEEEtai}

\usepackage[colorlinks,urlcolor=blue,linkcolor=blue,citecolor=blue]{hyperref}

\usepackage{color,array}

\usepackage{graphicx}
\usepackage{amsmath,amsfonts}
\usepackage{algorithmic}
\usepackage{algorithm}
\usepackage{textcomp}
\usepackage{stfloats}
\usepackage{url}
\usepackage{verbatim}
\usepackage{graphicx}
\usepackage{cite}
\usepackage{hyperref}  
\usepackage[table]{xcolor}
\usepackage{multirow}
\usepackage{booktabs}
\usepackage[most]{tcolorbox}

\begin{document}

\title{Robustness of Prompting: Enhancing Robustness of Large Language Models Against Prompt Attacks}

\author{Lin Mu, Guowei Chu, Li Ni, Lei Sang, Yiwen Zhang* 
\thanks{This work is supported by the National Natural Science Foundation of China( No.62206004, Grant No.62572002, Grant No.62272001, No.624065095); the Natural Science Foundation of Anhui Province (Grant No.2508085MF159, No.2308085MF213); the Hefei Key Technology R$\&$D "Champion-Based Selection" Project (Grant No.2024SGJ010). Corresponding author: Yiwen Zhang.}
\thanks{Lin Mu, Guowei Chu, Li Ni, Lei Sang and Yiwen Zhang are with the School of Computer Science and Technology, Anhui University, Hefei, Anhui, 230601, China. mulin@ahu.edu.cn, chuguowei@stu.ahu.edu.cn, nili@ahu.edu.cn, sanglei@ahu.edu.cn zhangyiwen@ahu.edu.cn.}
}

\markboth{Journal of IEEE Transactions on Artificial Intelligence, Vol. 00, No. 0, Month 2020}
{Lin Mu \MakeLowercase{\textit{et al.}}: Robustness of Prompting: Enhancing Robustness of Large Language Models Against Prompt Attacks}

\maketitle

\begin{abstract}
Large Language Models (LLMs) have demonstrated remarkable performance across various tasks by effectively utilizing a prompting strategy. However, they are highly sensitive to input perturbations, such as typographical errors or slight character order errors, which can significantly impair their performance. Despite advances in prompting techniques such as Chain-of-Thought and automatic prompt generation, developing a prompting strategy that explicitly mitigates the negative impact of such perturbations remains an open challenge.
To bridge this gap, we propose \textbf{R}obustness \textbf{o}f \textbf{P}rompting (\textbf{RoP}), a novel prompting strategy aimed at enhancing the robustness of LLMs. RoP consists of two stages: \textit{Error Correction} and \textit{Guidance}. In the \textit{Error Correction} stage, RoP applies diverse perturbation methods to generate adversarial examples, which are used to generate prompts that correct input errors automatically. In the \textit{Guidance} stage, RoP generates an optimal guidance prompt based on the corrected input, guiding the model to generate more robust and accurate inferences. Through comprehensive experiments spanning arithmetic, commonsense, and logical reasoning tasks, we demonstrate that RoP significantly improves LLMs' robustness against adversarial perturbations. Crucially, it preserves model accuracy with only minimal degradation compared to clean input scenarios, thereby establishing RoP as a practical and effective approach for enhancing LLM robustness in real-world applications.\footnote{Our code is available at \url{https://github.com/chuguowei/Robustness-of-Prompting}.}
\end{abstract}

\begin{IEEEImpStatement}
Robust prompt design is critical for the safe and reliable deployment of large language models (LLMs) in real-world applications. Despite impressive capabilities, current LLMs remain highly sensitive to minor prompt variations, which can lead to unstable outputs, unintended behaviors, or safety violations. This fragility limits their adoption in high-stakes domains such as healthcare, education, governance, and scientific decision support. The robustness-oriented prompt modeling and evaluation framework proposed in this work addresses these limitations by systematically identifying, measuring, and mitigating prompt-induced instability. By improving consistency, controllability, and reliability under prompt perturbations, our approach enables more dependable human–AI interaction and reduces the risk of misuse or misinterpretation. The resulting advances can support safer AI-assisted decision making, more trustworthy conversational agents, and more robust deployment of LLMs in socially sensitive and safety-critical environments, ultimately contributing to the responsible and sustainable integration of LLM technologies into society.
\end{IEEEImpStatement}


\section{Introduction}

\IEEEPARstart{L}{arge} Language Models (LLMs)~\cite{gpt3, chowdhery2023palm} have demonstrated remarkable capabilities across a wide range of tasks~\cite{wu2024survey, NetPrompt}. As the size of LLMs has increased, their in-context learning~\cite{gpt3} abilities have emerged. This allows them to observe the provided prompts, which include task-related examples or instructions, to solve the given input question without requiring any updates to their parameters. The design of prompting strategies is crucial, as it can significantly influence the performance of the LLMs in solving the given question. 

\begin{figure}[t]
  \centering
    \includegraphics[width=\linewidth]{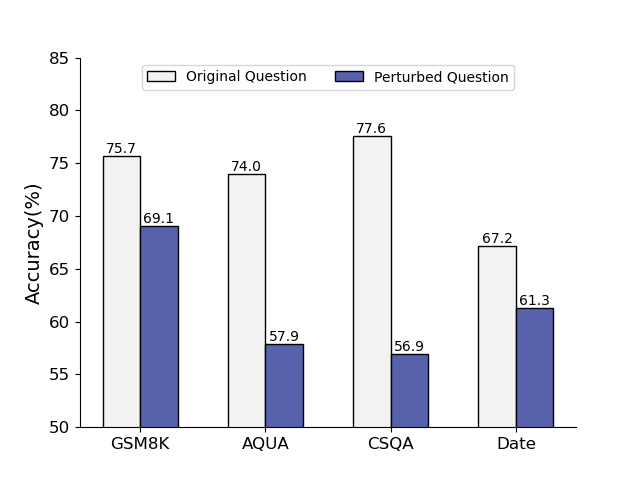}
    \caption{Comparison of the performance of GPT-3.5-Turbo before and after character-level perturbation in multiple reasoning tasks.}
  \label{fig:intro}
\end{figure}

\begin{figure*}[t]
  \centering
    \centering
    \includegraphics[width=0.75\textwidth]{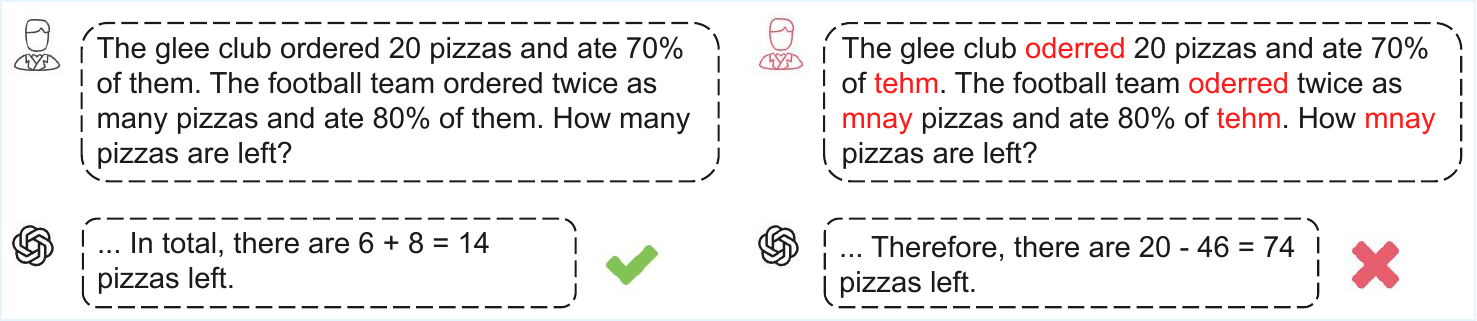}
    \caption{Subtle character-level changes in prompts can greatly affect the reasoning ability of LLMs. }
    \label{fig:cite}
\end{figure*}

Recent studies~\cite{wang2025multimodal} have explored various prompting strategies to further improve LLMs' performance. One notable method is Chain-of-Thought (CoT) prompting~\cite{Chain-of-thought}, which provides step-by-step reasoning examples to improve LLM output quality. However, CoT prompting requires careful manual design, which can be resource-intensive. In contrast, the Automatic Prompt Engineer (APE) method~\cite{APE} proposes an automatic generation and selection of prompts, eliminating manual intervention. Despite these advancements, a critical limitation persists: LLMs are highly sensitive to perturbations in input content~\cite{gan2024reasoning}. 
Even minor typographical mistakes or subtle character-level perturbations—as illustrated in Fig.~\ref{fig:cite}, such as changing “ordered” to “oderred”, “them” to “tehm”, or “many” to “mnay” in the input question—can significantly degrade the performance of LLMs. 
This fragility is illustrated in Fig.~\ref{fig:intro}, which shows a sharp decline in the inference accuracy of GPT-3.5-Turbo under minimal character-level perturbations across multiple reasoning tasks. While existing research highlights the sensitivity of LLMs to input content~\cite{xu2023llm, zhu2023promptrobust}, there remains a gap in understanding how to design a prompting strategy that enhances the robustness of these models against such perturbations.

To enhance the robustness of models, a widely used strategy is data augmentation~\cite{robustnessmodel}. This involves perturbing input texts and incorporating them into the training dataset to fine-tune the model. While data augmentation can help improve robustness, it requires updating model parameters, a process that can be both computationally expensive and resource-intensive.
To efficiently tackle this challenge, we propose a novel prompting strategy called \textbf{R}obustness \textbf{o}f \textbf{P}rompting (\textbf{RoP}), which aims at enhancing the robustness of LLMs without requiring updating their parameters. RoP introduces a two-stage pipeline comprising \textit{Error Correction} and \textit{Guidance}: In the \textit{Error Correction} stage, we generate a set of adversarial examples by applying various perturbation methods to input questions. These are paired with their corresponding unperturbed versions to automatically generate prompts that teach the LLMs to restore the original intended from perturbed inputs. In the \textit{Guidance} stage, the corrected input question is used to automatically generate an optimal guidance prompt that steers the LLM’s inference process, ensuring robust inference despite earlier input noise.

We evaluate RoP across three diverse reasoning task categories: arithmetic, commonsense, and logical. Experimental results demonstrate that while baseline large language models (LLMs) experience significant performance degradation under perturbed input conditions, RoP effectively mitigates this drop. In particular, when inputs are perturbed through the insertion of irrelevant or distracting information, RoP improves accuracy by an average of 16.1\% over baseline methods. These results underscore RoP’s effectiveness in preserving inference quality under adversarial conditions.

\section{Related Works}
\subsection{Automatic Prompting} 
Large Language Models (LLMs)~\cite{gpt3, chowdhery2023palm} have demonstrated substantial improvements in their ability to perform natural language reasoning tasks through a method known as prompting~\cite{gpt3, in-contextlearning, PromptLearning}. Prompting refers to the technique of guiding the LLMs, typically in the form of instructions or input-output pairs, that directs the model on how to solve a given task. To further boost LLM performance, Chain-of-Thought (CoT) prompting was introduced~\cite{Chain-of-thought}. This method includes intermediate reasoning steps within the prompt to guide the LLM through complex tasks. Subsequently, the concept of Zero-Shot-CoT was proposed~\cite{Zero-Shot-CoT}, where LLMs were able to perform reasoning tasks with minimal prior examples, by simply appending the trigger sentence \textit{Let's think step by step} to each question.

While these methods significantly improved LLM reasoning, they require pre-designed prompt settings, which may lead to suboptimal performance for certain tasks. In response, a growing body of research has focused on automating the process of prompt generation. The Automatic Prompt Engineer (APE)~\cite{APE} provides a promising solution by automatically generating prompts. It first creates a set of candidate instructions and then selects the best-performing one. Similarly, Optimization by PROmpting (OPRO)~\cite{OPRO} introduced an iterative optimization method that refines the prompt over multiple steps. In each iteration, the LLM generates new candidate solutions based on the current prompt, incorporating previously generated solutions and their corresponding evaluation values. These refined solutions are then used to further optimize the prompt.
\begin{figure*}[t]
  \centering
    \centering
    \includegraphics[width=\linewidth]{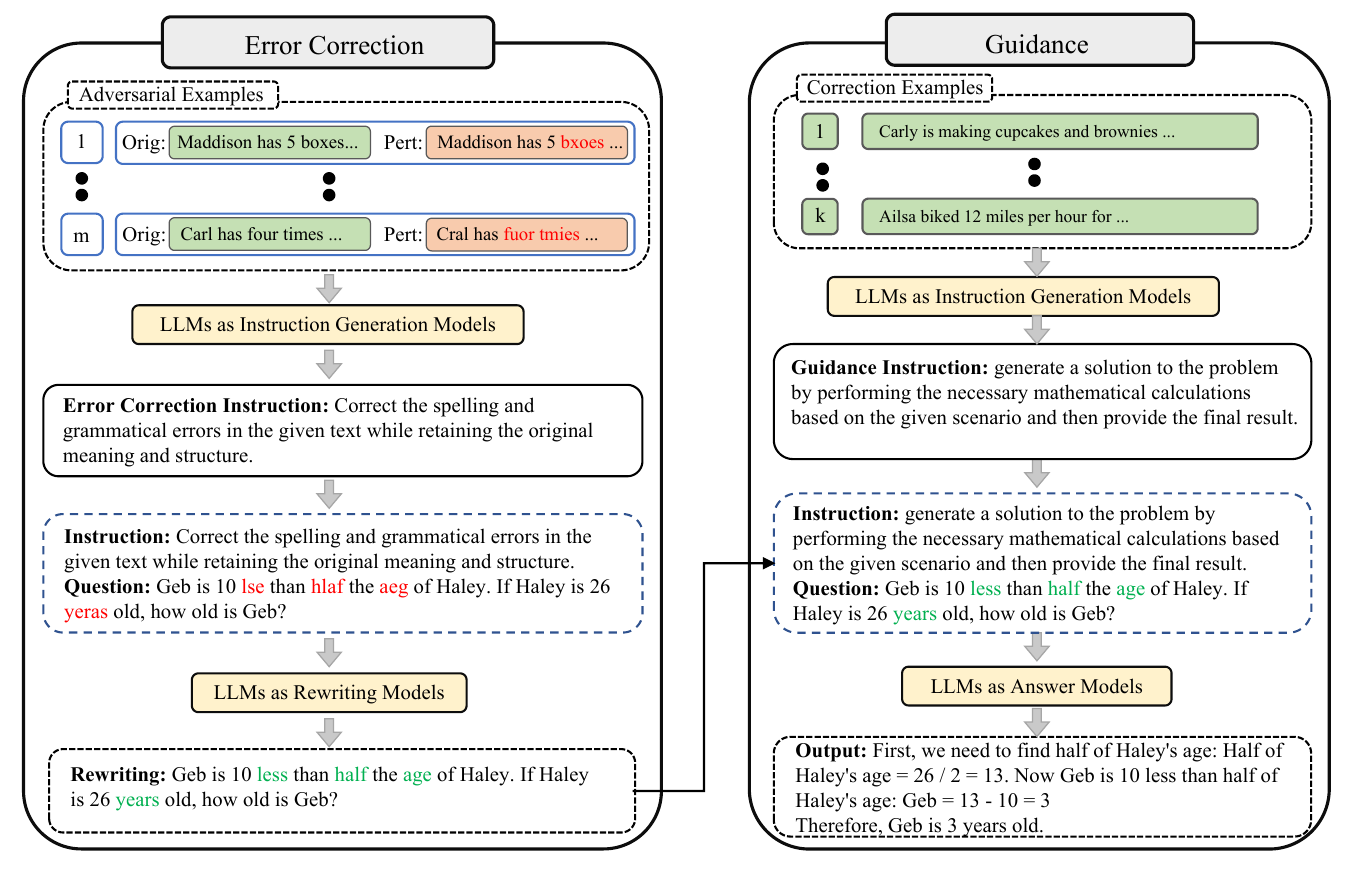}
    \caption{The framework of Robustness of Prompting (RoP). RoP consists of two stages: \textit{Error Correction} and \textit{Guidance}. In the \textit{Error Correction} stage, RoP generates adversarial examples to automatically generate prompts that teach the LLMs to correct the inputs. In the \textit{Guidance} stage, the corrected input question is used to automatically generate an optimal guidance prompting that steers the LLM’s inference process.}
    \label{fig:framework}
\end{figure*}

\subsection{Adversarial Attack}
Despite the advancements in automatic prompting techniques, LLMs remain highly sensitive to the input text. Even the smallest perturbation in the input can lead to a substantial degradation in their performance.  For example, research by~\cite{gan2024reasoning} demonstrated that a single character-level perturbation in an input question can significantly affect inference accuracy. As reported in their study, the accuracy of the Mistral-7B-Instructs model~\footnote{https://huggingface.co/mistralai/Mistral-7B-Instruct-v0.2} dropped by 5\% on the GSM8K~\cite{gsm8k} dataset due to such perturbations.
Moreover, recent work by~\cite{xu2023llm} explored adversarial attacks targeting the input, further confirming LLMs' vulnerability to even minor changes in the text. These perturbations can range from character-level errors to more complex changes in words, sentences, or semantics. For instance,~\cite{zhu2023promptrobust} proposed methods for adversarial attacks at multiple levels: character, word, sentence, and semantic, highlighting the extent of the issue across different types of input modifications.

While existing research has examined on various adversarial attack techniques, there remains a significant gap in efforts to enhance the robustness of LLMs against such perturbed inputs. The challenge of designing effective robustness prompting strategies—particularly in response to perturbed text—has not been thoroughly explored. Therefore, this paper aims to investigate methods for enhancing the robustness of LLMs against question-based perturbations.

\section{Method}
LLMs are highly sensitive to input perturbations, where even minor perturbations in the question can significantly impact their output. Given these vulnerabilities, we propose the Robustness of Prompting (RoP) strategy to enhance the robustness of LLM responses. The overall framework is illustrated in Fig.~\ref{fig:framework}.

\subsection{Background}
The paradigm of in-context learning~\cite{incontextlearning} allows LLMs to perform tasks by using only a few examples and an instruction within a prompt. Formally, given a question $x$ and a set of candidate answers $Y=\{y_1,\dots,y_n\}$, an LLM, denoted as $L_\theta$ and parameterized by $\theta$, predicts the answer that maximizes the conditional probability among the candidate answers based on the provided prompt $P$. The prompt $P$ consists of an instruction and $k$ examples, represented as $P= <in, \{e(x_1, y_1),\dots,e(x_k, y_k)\}>$, where $in$ denotes the description and requirements of the task and $e(x_i, y_i)$ denotes an example expressed in natural language. Here, $x_i$ denotes the demonstration question, and $y_i$ is its corresponding answer. Accordingly, the in-context learning formulation can be expressed as follows:
\begin{equation}
\begin{aligned}
    \hat{y} = &\mathop{\arg\max}\limits_{y_i \in Y} L_{\theta}(y \mid x, P) \\
            = &\mathop{\arg\max}\limits_{y_i \in Y} L_{\theta}(y \mid x, in, e(x_1, y_1), \ldots, e(x_k, y_k))
\end{aligned}
\label{equ:incontext}
\end{equation}

While in-context learning enables LLMs to perform question based prompts, this framework assumes ideal conditions. However, input noise and inconsistencies are inevitable, necessitating robust mechanisms to preserve model reliability under perturbation.

\subsection{Perturbation Methods}\label{sec:pertubation}
To evaluate and enhance the robustness of LLMs, we introduce five distinct types of perturbations that simulate common errors in real-world inputs. These perturbations serve dual purposes: assessing model fragility and constructing adversarial examples that facilitate error correction instruction generation. The five designed perturbation types are as follows:

\begin{itemize}
    \item Error Character (EC): This type of perturbation involves randomly shuffling the internal characters of words or changing characters to other (e.g., \textit{times} $\rightarrow$ \textit{tmies}, \textit{will} $\rightarrow$ \textit{wlil}).
    \item Similar Character (SC): This type of perturbation involves replacing one or more characters with visually similar symbols or special characters (e.g., \textit{will} $\rightarrow$ \textit{wiļļ}, \textit{times} $\rightarrow$ \textit{tīmês}).
    \item Words Out of Order (WOO): This type of perturbation involves rearranging neighboring word positions to disrupt syntactic structure (e.g., \textit{6 times older} $\rightarrow$ \textit{older 6 times}, \textit{3 times}$\rightarrow$ \textit{times 3}).
    \item Homophone Words (HW): This type of perturbation involves replacing original terms with phonetically equivalent alternatives (e.g., \textit{be} $\rightarrow$ \textit{bee}).
    \item Unaffected Interference Conditions (UIC): This type of perturbation involves appending irrelevant but plausible information that does not alter the answer (e.g., \textit{Ruby is 6 times older than Sam.} $\rightarrow$ \textit{Ruby is 6 times older than Sam. Nine years ago, Ruby bought a clock that cost \$45, and she plans to replace it in 7 years.}).
\end{itemize}

In this work, we use the term ‘adversarial examples’ to refer to input questions that have been perturbed using our proposed methods (EC, SC, WOO, HW, UIC), with the goal of simulating real-world noise and errors. While not all such perturbations necessarily cause the model to fail, they represent a diverse set of challenges that can degrade model performance.

To automate the generation of perturbed questions $\hat{x}$ from an original question $x$ and its answer $y$, we employ the LLM itself as a perturbation model. Given a perturbation type $pt$, we define a perturbation prompt $ap$ as follows:

\noindent\rule{\linewidth}{1pt}

\noindent \textit{Your objective is to rewrite a given math question using the following perturbation strategy. The rewritten question should be reasonable, understandable, and able to be responded to by humans.}\\
\textit{Perturbation type: $pt$}\\
\textit{The given question: \{$x$\}}\\
\textit{Answer of the given question: \{$y$\}} \\
\textit{Please rewrite the question using the specified perturbation strategy, and avoid significant deviation in the question content.\\
It is important to ensure that the rewritten question has only one required numerical answer. You just need to print the rewritten question without answer.}

\noindent\rule{\linewidth}{1pt}

\begin{figure}[t]
  \centering
    \centering
    \includegraphics[width=\linewidth]{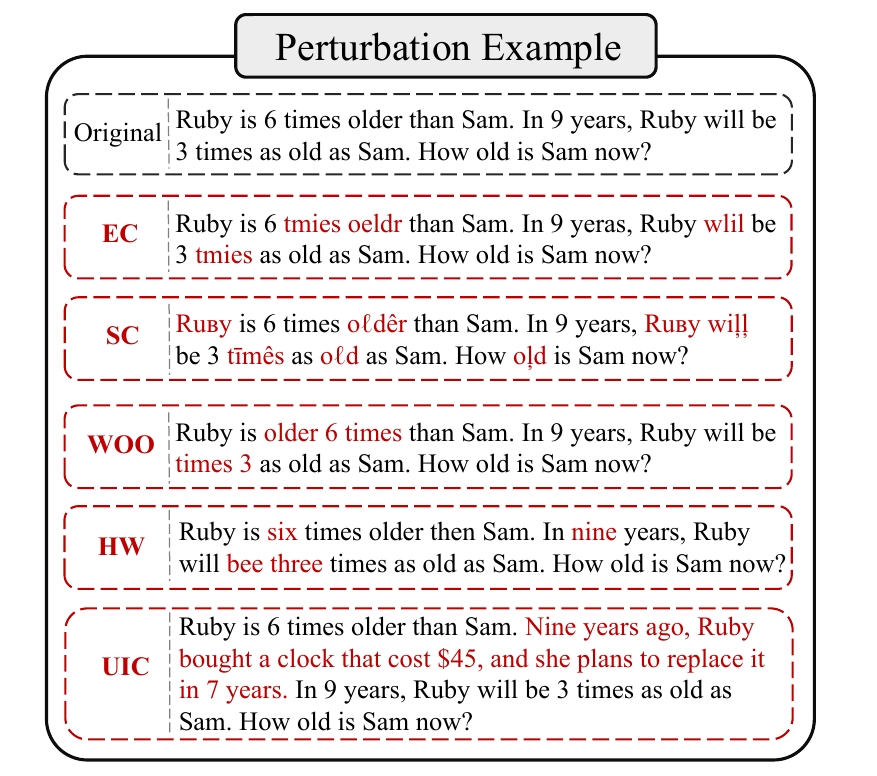}
    \caption{Five different perturbation methods.}
    \label{fig: example}
\end{figure}

With the perturbation prompt, we can generate a perturbation question $\hat{x}$ based on $x$. Fig.~\ref{fig: example} shows the five different perturbation methods based on a given question.

\subsection{Error Correction}
Building on the adversarial examples generated via perturbation, the first stage of RoP—Error Correction—is designed to recover the intended semantics of flawed inputs through correctional prompts.

\textbf{Adversarial sample Generation}: Inspired by adversarial attack in other fields~\cite{robustnessmodel, wang2023adversarial}, we synthesize adversarial examples from perturbed inputs. Specifically, we sample $k$ questions from the training dataset and apply aforementioned perturbation methods to obtain pairs $<x_i, \hat{x_i}> \in \mathcal{D}_{adv}$, where $x_i$ is the original question, and $\hat{x_i}$ its perturbed variant. 

\textbf{Error Correction Instruction}: The next step is to convert these adversarial examples into a prompt-based correction task. To this end, we sample a subset of $m$ such examples, denoted as $\hat{\mathcal{D}}_{adv} \subseteq \mathcal{D}_{adv}$, and use the APE framework~\cite{APE} to automatically generate an instruction prompt $in_{ec}$. For the subset samples $<x_m, \hat{x_m}> \in \hat{\mathcal{D}}_{adv}$, we define the meta-prompt for automatically generating the instruction prompt $in_{ec}$ as follows:

\noindent\rule{\linewidth}{1pt}

\noindent \textit{You are a kick ass prompt engineer, you are given with input variables and the output generated by an LLM. Find the right prompt for this batch of input and outputs.}\\
\textit{Input: $\hat{x_1}$}\\
\textit{Output: $x_1$}\\
...\\
\textit{Input: $\hat{x_m}$}\\
\textit{Output: $x_m$}

\noindent\rule{\linewidth}{1pt}

The design of $in_{ec}$ aims to instruct the LLM to recognize and correct perturbations in input questions while preserving the answer. The final error-corrected prompt is thus defined as: $P=<in_{ec}, \hat{x} >$, which is passed to the LLM along with a perturbed question $\hat{x}$. The model then outputs a corrected version $\hat{x}_{ec}$ that aligns semantically with the original input $x$.

\begin{table*}[t]
    \centering
    \caption{Accuracy(\%) comparison of five perturbation methods with various prompting strategies on arithmetic reasoning tasks. No Dert.: indicates that the input questions have not been perturbed. $\Downarrow_{(*)}$: indicates that accuracy decreases compared to the Stand with No Dert. input. }
    \label{tab:mainresult}
    \setlength{\tabcolsep}{12pt}
    \resizebox{\linewidth}{!}{
    \begin{tabular}{cl|cccccc|c}
        \toprule
        \textbf{Perturb}& \textbf{Method} & \textbf{SingleEq}  & \textbf{AddSub}  & \textbf{MultiArith}  & \textbf{AQUA}  &   \textbf{GSM8K} & \textbf{SVAMP} & \textbf{Avg.}\\
        \midrule
        No Dert.& Stand   & 95.7 & 90.4 & 90.5  & 74.0  &  75.7 & 79.4 & 84.3 \\
        \midrule
        \multirow{5}{*}{EC}  & Stand  & 89.2 & 86.8 &  87.0 &  57.9 &  69.0 &75.0  & 77.5 $(\Downarrow_{6.8})$ \\  
        & CoT &90.9&	85.3&	\textbf{96.3}&	57.9&	70.2&	\textbf{80.3} & 80.2 $(\Downarrow_{4.1})$ \\
        &APE   &90.9&	76.7&	91.2&	57.9&	66.9&	76.5 & 76.7 $(\Downarrow_{7.6})$\\
        & PromptAgent   & 92.5&	80.8&	93.5&	56.3&	76.5&	76.5 & 79.3 $(\Downarrow_{5.0})$\\
        & \textbf{RoP (Our) }  & \textbf{95.3}&	\textbf{87.3}&	95.0&	\textbf{63.0}&	\textbf{72.9}&	79.7 & \textbf{82.2} $(\Downarrow_{\textbf{2.1}})$\\
        \midrule
        \multirow{5}{*}{SC}  & Stand  &88.0	&88.1 &	85.3&	52.4&	68.0&	74.0 & 76.0 $(\Downarrow_{8.3})$\\ 
        &CoT & 88.4 & 86.3&	\textbf{94.2}& 	65.8&	74.2&	76.2&	80.8  $(\Downarrow_{3.5})$\\
        &APE  & 92.7	& \textbf{90.6}&	86.2&	52.8	&71.4&	76.2 & 78.3  $(\Downarrow_{6.0})$\\
        & PromptAgent   &92.5 &46.3&	85.0&	58.7&	67.3&	77.4 & 71.2 $(\Downarrow_{13.1})$\\
        & \textbf{RoP (Our)}   &\textbf{95.1} &	82.5&	92.3&	\textbf{70.1}&	\textbf{75.4}	&\textbf{78.3}& \textbf{82.3}  $(\Downarrow_{\textbf{2.0}})$\\
        \midrule
        \multirow{5}{*}{WOO} & Stand & 88.8&	88.6&	86.8&	59.5&	70.1&	76.9  & 78.4 $(\Downarrow_{5.9})$\\ 
        &CoT & 88.2 &	86.3& 	\textbf{96.5}&	56.7 & \textbf{73.2}&	78.2&	79.9  $(\Downarrow_{4.4})$ \\
        &APE  &\textbf{92.3} &	\textbf{91.7} &	88.0&	60.6&	71.3&	72.7 & 79.4 $(\Downarrow_{4.9})$\\
        & PromptAgent  &85.6&	83.5&	88.8&	57.1&	71.4&	76.8 &77.2 $(\Downarrow_{7.1})$\\
        & \textbf{RoP (Our)}  &89.0&	84.6&	91.3&	\textbf{65.8}&	71.3&	\textbf{79.5} & \textbf{80.2} $(\Downarrow_{\textbf{4.1}})$\\
        \midrule
         \multirow{5}{*}{HW} & Stand & 84.1&	82.0&	78.7&	56.3 & 66.4&	65.9 &72.2 $(\Downarrow_{12.1})$\\ 
         & CoT & 87.4& 	84.3&	92.7& 63.4 & 70.4 &	75.8 &	79.0 $(\Downarrow_{5.3})$\\
        &APE  &83.1&	82.8&	89.3&	59.8&	68.5&	77.8 &76.9 $(\Downarrow_{7.4})$\\
        & PromptAgent & 87.8&	83.3&	82.8&	58.3	&66.5&	75.5 &75.7 $(\Downarrow_{8.6})$ \\
        & \textbf{RoP (Our)}  &\textbf{90.9}	&\textbf{84.3}&	\textbf{94.0}&	\textbf{64.6}&	\textbf{74.2}&	\textbf{82.4} & \textbf{81.7} $(\Downarrow_{\textbf{2.6}})$\\
        \midrule
        \multirow{5}{*}{UIC}  & Stand &63.6&	72.2&	54.5&	57.1&	51.9&	54.2 &58.9  $(\Downarrow_{25.4})$\\
       & CoT   &   62.2&	71.7& 59.3&	50.0&	51.4&	55.7 & 58.4 $(\Downarrow_{25.9})$\\
        &APE   &	78.7& 69.9& 59.0&	56.3&	45.0&	65.5 & 62.4 $(\Downarrow_{21.9})$\\
        & PromptAgent   &	57.3& 57.0&	50.5&	48.4&	49.8&	49.2 & 52.0 $(\Downarrow_{32.3})$\\
        & \textbf{RoP (Our) }  & \textbf{79.1}&	\textbf{89.9}&	\textbf{74.7}&	\textbf{62.2}&	\textbf{61.7}&	\textbf{76.7} & \textbf{74.0} $(\Downarrow_{\textbf{10.3}})$\\
        \bottomrule
    \end{tabular}}
    
\end{table*}

\textbf{Execute Correction}: The instruction $in_{ec}$ and the adversarial subset $\hat{\mathcal{D}}_{adv}$ are provided to an LLM, such as GPT-4o. The model utilizes the prompt $P = <in_{ec}, \hat{x}>$ to rewrite an erroneous input $\hat{x}$, producing a corrected version $\hat{x}_{ec}$ that restores the original semantics despite the presence of input perturbations.

\subsection{Guidance Instruction Generation}
In the previous stage, we generated corrected questions based on the error correction instruction $in_{ec}$. To automatically generate optimal prompts for these corrected questions, we use APE~\cite{APE} to generate optimized instructions $in_{opt}$. These instructions guide the LLMs in producing robust responses using adversarial samples. Specifically, we randomly sample $k$ question-answer pairs $<\hat{x}_{ec}^i, a^i> \in \mathcal{D}_{cor}$. The guidance instruction $in_{opt}$ generate using APE. For $\mathcal{D}_{cor}$, we define the meta-prompt for automatically generating the instruction prompt $in_{opt}$ as follows:

\noindent\rule{\linewidth}{1pt}

\noindent \textit{You are a kick ass prompt engineer, you are given with input variables and the output generated by an LLM. Find the right prompt for this batch of input and outputs.}\\
\textit{Input: $\hat{x}_{ec}^1$}\\
\textit{Output: $a^1$}\\
...\\
\textit{Input: $\hat{x}_{ec}^i$}\\
\textit{Output: $a^i$}

\noindent\rule{\linewidth}{1pt}

Finally, given a question $x_{ec}$, the answer $\hat{y}_{ec}$ is generated:
\begin{equation}
\begin{aligned}
\hat{y}_{ec} = & \mathop{\arg \max}\limits_{y_i \in Y} L_{\theta}(y \mid x_{ec}, in_{opt}, \mathcal{D}_{cor})  \\
        = &\mathop{\arg \max}\limits_{y_i \in Y} L_{\theta}(y \mid x_{ec}, in_{opt}, <\hat{x}_{ec}^1, a^1> \\ & \dots <\hat{x}_{ec}^k, a^k>
\end{aligned}
\label{equ:final}
\end{equation}

\subsection{Synergistic Effect}

The superior performance of RoP stems from a synergistic interaction between the Error Correction and Guidance stages, rather than the effect of either component alone. This synergy arises from two key mechanisms: distribution alignment and task decomposition.

When inputs are subjected to perturbations, their distribution deviates significantly from the clean data distribution under which prompts are typically optimized. As described in Eq. (1), the model $L_\theta$ generates predictions by maximizing the conditional probability given the prompt $P$. However, when the input $x$ is perturbed into $\hat{x}$, this distribution shift degrades the effectiveness of both instruction following and in-context learning.

A single-stage prompting strategy must simultaneously (i) recover the original semantics from $\hat{x}$ and (ii) perform multi-step reasoning. This creates a capacity conflict, as the model is required to solve two fundamentally different objectives within a single prompt, which empirically leads to suboptimal performance.

To address this limitation, RoP adopts a two-stage modular design.

\textbf{Error Correction Stage.}
Given adversarial pairs $\langle \hat{x}, x \rangle \in \hat{D}_{adv}$, the model generates a corrected input:
\begin{equation}
\begin{aligned}
\hat{x}_{ec} = L_\theta(\hat{x} \mid in_{ec}),
\end{aligned}
\label{equ:incontext3}
\end{equation}
where $\hat{x}_{ec} \approx x$. This stage reduces surface-level perturbations and produces a semantically consistent input, effectively stabilizing the input distribution.

\textbf{Guidance Stage.}
Based on the corrected samples $D_{cor} = {\langle \hat{x}{ec}^i, a^i \rangle}$, the model optimizes the guidance instruction $in{opt}$ and performs inference:
\begin{equation}
\begin{aligned}
\hat{y} = \arg\max_{y_i \in Y} L_\theta(y \mid \hat{x}_{ec}, in_{opt}, D_{cor}).
\end{aligned}
\label{equ:incontext4}
\end{equation}

Since both the query input and in-context examples are derived from $\hat{x}_{ec}$, they share a consistent distribution, which is crucial for effective in-context learning.

The synergistic effect can be understood by comparing different configurations:
1) Correction Only improves input quality but lacks task-specific reasoning optimization.
2) Guidance Only optimizes reasoning but operates on perturbed inputs, leading to a mismatch between the input and the prompt distribution.
3) RoP (Combined) first aligns the input distribution and then optimizes reasoning on this stabilized space, ensuring consistency between the query and demonstration examples.

As a result, the Guidance stage can learn prompt templates tailored to high-fidelity reasoning paths, which would not be discoverable under noisy inputs. Therefore, the performance gain of RoP originates from the interaction between distribution correction and instruction optimization, which cannot be achieved by either stage alone or by a single-stage prompting framework.

\section{Experiments}\label{experiment}


\subsection{Arithmetic Reasoning}
\subsubsection{Experiment Setup}
In this section, we evaluate the effectiveness and performance of our method, RoP, across six arithmetic reasoning datasets. These datasets are:
\begin{itemize}
    \item \textbf{GSM8K}~\cite{gsm8k} is a dataset consisting of 8,500 high-quality, linguistically diverse elementary math problems created by humans. Of these, 7,500 problems are designated for training, while the remaining 1,000 are reserved for testing. These math problems typically require between 2 and 8 steps to solve and involve fundamental arithmetic operations, including addition, subtraction, multiplication, and division. In this study, we evaluate our method using 1,319 test questions.
    \item \textbf{AQUA-RAT}~\cite{aqua} is a dataset comprising 100,000 multiple-choice questions focused on mathematics, encompassing a wide range of topics and varying levels of difficulty. It is regarded as the most challenging dataset among the six arithmetic reasoning datasets. For our evaluation, we use a carefully curated set of 250 test questions.
    \item \textbf{SingleEq} (Single Equation) ~\cite{SingleEq} is a dataset containing 508 elementary school algebra word problems, each of which can be represented by a single equation involving multiple arithmetic operations.
    \item \textbf{SVAMP} (Simple Variations on Arithmetic Math Word Problems)~\cite{SVAMP} is a dataset that comprises arithmetic word problems appropriate for fourth-grade students and below. It contains a total of 1,000 test questions. 
    \item \textbf{MultiArith}~\cite{Multiarith} is a dataset that includes a collection of polynomial arithmetic problems, consisting of a series of questions and corresponding answers. This dataset is commonly used for training and evaluating the performance of machine learning models for polynomial arithmetic. 
    \item \textbf{AddSub}~\cite{Addsub} is a dataset consisting of 395 problems specifically focused on addition and subtraction. It focuses on evaluating the model's fundamental ability to handle the core additive and subtractive logic within textual contexts.
\end{itemize}

To evaluate the robustness of RoP against question perturbations, we apply five types of input perturbations described in Section Method to the test dataset. Specifically, we use GPT-4o~\cite{gpt-4o} to generate the perturbed data, along with the error correction instructions and the guidance instructions required for the experiment. For the evaluation phase, we primarily employ GPT-3.5-turbo~\cite{gpt-3.5-turbo} to assess the performance of each method across various datasets. Additionally, to further evaluate the scalability of RoP, we conducted experiments using three more evaluation models: GPT-4o~\cite{gpt-4o}, o1-mini~\cite{o1-mini}, and o3-mini~\cite{o3-mini}~\footnote{https://openai.com}.


\begin{itemize}
    \item {Standard Prompting (Stand)}: The original unperturbed or perturbed question is directly fed into the model without any enhancements.
    \item {Chain of Thought (CoT)}~\cite{Chain-of-thought}: Prompts include exemplars of intermediate reasoning steps to improve model inference on multi-step problems. 
    \item {Automatic Prompt Engineer (APE)}~\cite{APE}: An automated method that generates high-quality task instructions via prompt selection and optimization.
    \item {PromptAgent}~\cite{promptagent}: A state-of-the-art agent-based prompt optimization framework that autonomously constructs expert-level prompts through iterative refinement.
\end{itemize}

\begin{table}[t]
    \centering
    \caption{The result of the ablation study.}
    \label{tab:abation}
    \begin{tabular}{ccccc}
        \toprule
        \textbf{Perturb} & \textbf{Method} & \textbf{AQUA}  &  \textbf{GSM8K} & \textbf{SingleEq} \\
        \midrule
        No Dert. & Stand & 74.0 & 75.7 & 95.7 \\
        \midrule
        \multirow{4}{*}{EC} & Stand & 57.9 & 69.1 & 89.2 \\
        &CO   & 61.0 & 73.8 & 94.7 \\
        &GO  & 57.9 & 66.9 & 90.9 \\
        &Direct  & 53.2 & 69.1 & 83.7 \\
        &\textbf{RoP}  & \textbf{63.0} &	\textbf{73.9} & \textbf{95.3} \\
        \midrule
        \multirow{4}{*}{HW} & Stand & 56.3 & 66.4 &	84.1 \\
        & CO & 59.1  & 73.2 & 87.4 \\
        & GO & 59.8  & 68.5  & 83.1 \\
        &Direct  & 62.2 & 66.2 & 80.7 \\
        & \textbf{RoP} & \textbf{64.6} & \textbf{74.2} & \textbf{90.9} \\
        \bottomrule
    \end{tabular}
\end{table}

\subsubsection{Result}
In this section, we compare the RoP method with the baseline approaches—Stand, CoT, APE, and PromptAgent—across six arithmetic reasoning datasets under various textual perturbations. First, we verify the inference accuracy of LLMs using Equation~\ref{equ:incontext}, applied to the original, unperturbed datasets. Then, we evaluate the inference accuracy of LLMs after applying each baseline method to the perturbed datasets. The result, summarized in Table~\ref{tab:mainresult}. The findings highlight that RoP demonstrates superior robustness across most perturbation scenarios. Notably, for complex arithmetic reasoning tasks such as AQUA, GSM8K, and SVAMP, RoP maintains a relatively high level of accuracy, with only marginal reductions in performance following perturbation. This contrasts with baseline methods, which show more significant drops in accuracy.

In particular, the UIC perturbation posed a considerable challenge to the LLMs. The inference accuracy dropped from 84.3\% on the clean dataset to 58.9\% post-perturbation, a decrease of 25.4\%. While traditional baseline approaches failed to mitigate this drop effectively, the RoP method limited the performance degradation to just 10.3\%, thereby illustrating its capacity to enhance model robustness under semantic perturbations. However, under the WOO perturbation scenario, the advantage of RoP was less pronounced. In this case, the results are similar to those of CoT and APE. This outcome is likely due to the inherent ability of LLMs to handle minor syntactic reordering, given their training on diverse and syntactically varied corpora.


\begin{table}[t]
    \centering
    \caption{Accuracy(\%) of different LLM on AQUA dataset.}
    \label{tab:othermodel}
\begin{tabular}{ll|ccc|c}
\toprule
\textbf{Perturb} & \textbf{Method} & \textbf{GPT-4o} & \textbf{o1-mini} & \textbf{o3-mini} & \textbf{Avg.} \\
\midrule
No Dert. & Stand & 79.53 & 83.46 & 85.43 & 82.81 \\
\midrule
 \multirow{5}{*}{EC}
& Stand & 79.13 & 80.71 & 84.65 & 81.50 \\
 & CoT & 70.87 & 81.89 & 79.53 & 77.43 \\
 & APE & 80.31 & 79.53 & 85.43 & 81.76 \\
 & PromptAgent & 80.71 & 81.50 & \textbf{85.83} & 82.68 \\
 & \textbf{RoP (Our)} & \textbf{81.50} & \textbf{82.68} & 84.25 & \textbf{82.81} \\
 \midrule
 \multirow{5}{*}{SC}
 & Stand & 71.65 & \textbf{87.80} & 82.68 & 80.71 \\
 & CoT & 77.95 & 84.25 & 84.65 & 82.28 \\
 & APE & 74.41 & 70.87 & 85.83 & 77.04 \\
 & PromptAgent & \textbf{77.95} & 84.65 & 83.07 & 81.89 \\
 & \textbf{RoP (Our)} & 75.20 & 85.43 & \textbf{87.40} & \textbf{82.68} \\
\midrule
 \multirow{5}{*}{WOO}
 & Stand & 78.35 & 81.89 & 81.50 & 80.58 \\
 & CoT & 79.13 & 81.50 & 84.25 & 81.63 \\
 & APE & \textbf{79.92} & 72.44 & 72.83 & 75.06 \\
 & PromptAgent & 79.13 & 85.83 & 85.43 & 83.46 \\
 & \textbf{RoP (Our)} & 79.13 & \textbf{86.22} & \textbf{87.80} & \textbf{84.38} \\
\midrule
\multirow{5}{*}{HW}
& Stand & 77.56 & 82.28 & 85.83 & 81.89 \\
& CoT & 81.10 & 83.46 & \textbf{86.61} & 83.72 \\
& APE & 79.92 & 42.52 & 83.07 & 68.50 \\
& PromptAgent & 77.17 & 82.68 & 85.83 & 81.89 \\
& \textbf{RoP (Our)} & \textbf{81.10} & \textbf{85.83} & 85.83 & \textbf{84.25} \\
 \midrule
 \multirow{5}{*}{UIC}
& Stand & 70.87 & 80.71 & 80.71 & 77.43 \\
& CoT & 74.41 & \textbf{82.68} & 81.10 & 79.40 \\
& APE & 72.44 & 48.03 & 50.79 & 57.09 \\
& PromptAgent & 73.23 & 81.10 & 79.92 & 78.08 \\
& \textbf{RoP (Our)} & \textbf{77.95} & 80.31 & \textbf{81.50} & \textbf{79.92} \\
\bottomrule
    \end{tabular}
\end{table}

\subsubsection{Ablation Study}
To systematically evaluate the impact of \textit{Error Correction} and \textit{Guidance} components of RoP, we conducted an ablation study. 
 This study compares three settings: (1) Correction Only (CO), where only the \textit{Error Correction} module is used without the \textit{Guidance} module. (2) Guidance Only (GO), where only the \textit{Guidance} module is used without the \textit{Error Correction} (essentially the APE method). (3) Directly correct errors and provide answers (Direct),  where the model is instructed to directly correct errors and provide answers without our specific modular framework. For this experiment, GPT-3.5-Turbo was used as the evaluation model. 

As shown in Table~\ref{tab:abation}, we present the experiment results of two perturbation methods (EC and HW), applied to AQUA, GSM8K and SingleEq. The findings reveal that using only the \textit{Error Correction} module results in a smaller decrease in LLM's performance, suggesting that \textit{Error Correction} alone can effectively enhance the robustness of the LLMs. Similarly, the \textit{Guidance} module also enhances the robustness of the LLMs when used alone. Notably, the combined use of both components (RoP) consistently achieved superior performance across both datasets and perturbation types. This demonstrates that the synergy between \textit{Error Correction} and \textit{Guidance} significantly amplifies the robustness of LLMs when dealing with input perturbations. This underscores the necessity of both components working together to optimize LLMs' performance under perturbed conditions. The method of directly correcting errors and providing answers performs poorly, demonstrating performance that is even lower than the standard baseline (Stand) across most tested datasets and perturbation types. This indicates that LLMs are not adept at directly handling input perturbations while generating answers.

\begin{figure*}
  \centering
  \begin{minipage}[t]{0.32\linewidth}
    \centering
    \includegraphics[width=\linewidth]{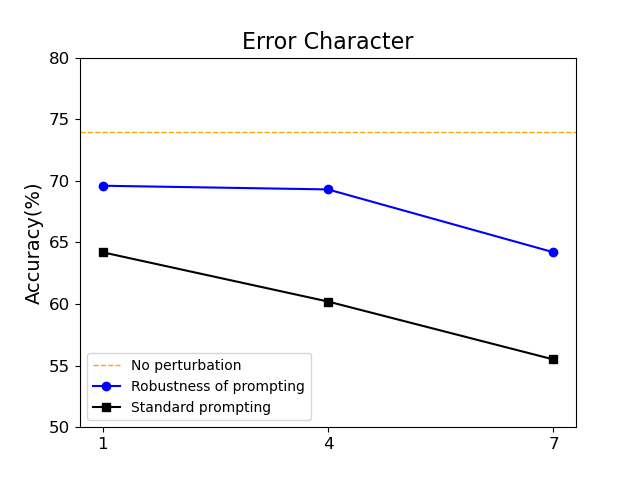}
    {(a) EC}
  \end{minipage}
  \begin{minipage}[t]{0.32\linewidth}
    \centering
    \includegraphics[width=\linewidth]{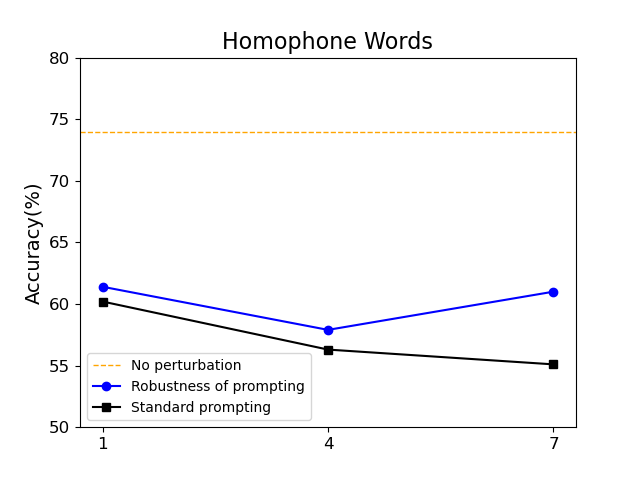}
    {(b) HW}
  \end{minipage}
    \begin{minipage}[t]{0.32\linewidth}
    \centering
    \includegraphics[width=\linewidth]{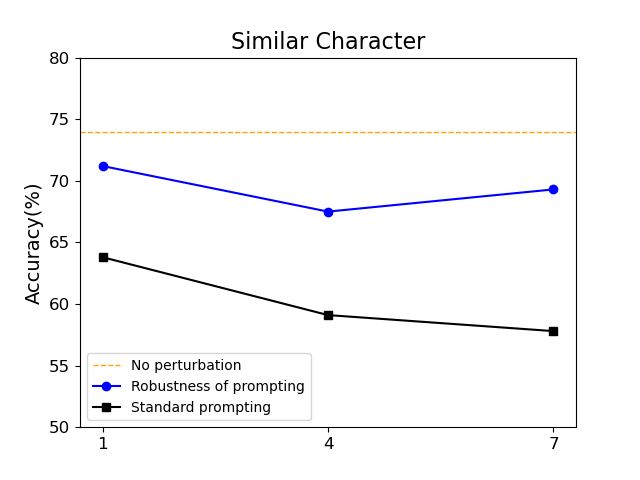}
    {(c) SC}
  \end{minipage}
  \caption{The effectiveness of different levels of perturbation.}
  \label{fig: levelex}
\end{figure*}


\subsubsection{Performance on Other Models} 
To further explore the scalability and cross-model generalizability of the RoP, we conducted a comprehensive evaluation using the AQUA dataset and a variety of LLMs. This investigation sought to determine whether RoP maintains its robustness-enhancing properties when applied to different model architectures and capacities. We tested RoP on more advanced and lightweight models, including GPT-4o, o1-mini, and o3-mini. Notably, GPT-4o was consistently employed as the optimizer model, ensuring consistency in optimization dynamics across experimental conditions. 

The results of these experiments, which used various LLMs and two perturbation methods, are presented in Table~\ref{tab:othermodel}. The result shown that RoP demonstrated strong transferability, consistently mitigating performance degradation caused by perturbations across all evaluated LLMs.

\subsubsection{Robustness Across Perturbation Levels}
To assess the granularity of RoP’s robustness under increasing textual interference, we conducted a controlled experiment wherein the number of perturbed characters in each input question was systematically varied—specifically at 1, 4, and 7-character perturbation levels. This test evaluates the model's ability to maintain inference accuracy as perturbation severity increases. As shown in Fig.~\ref{fig: levelex}, RoP consistently improves accuracy across all levels of perturbation. Although performance degradation is observed as perturbation intensity increases (i.e., from 1 to 7 characters), the decline in accuracy for RoP is significantly less steep than that of unoptimized prompting strategies. This indicates that RoP effectively maintains strong performance, even when faced with various forms of textual interference. These findings underscore the gradual robustness of RoP, confirming that while no method remains entirely immune to aggressive perturbations, RoP degrades more gracefully, sustaining competitive accuracy under both mild and severe distortions. This robustness to variable perturbation intensity is a critical feature for real-world deployment in noisy or adversarial environments.

\begin{figure*}[t]
  \centering
  \begin{minipage}[b]{0.43\linewidth}
    \centering
    \includegraphics[width=\linewidth]{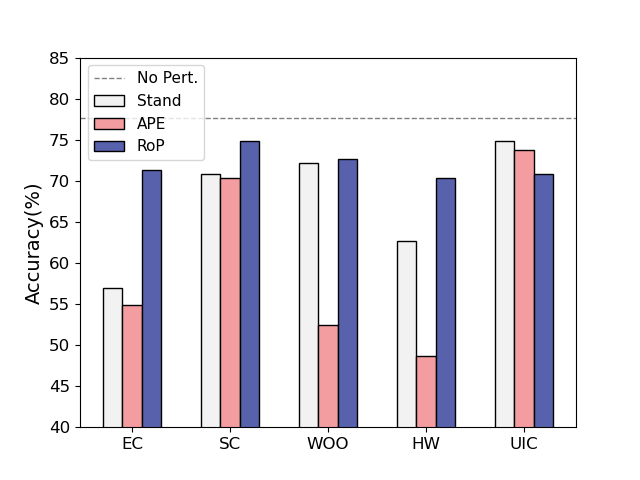}
    {(a) CSQA}
    \label{fig:subfig-CSQA}
  \end{minipage}
  \hfill
  \begin{minipage}[b]{0.43\linewidth}
    \centering
    \includegraphics[width=\linewidth]{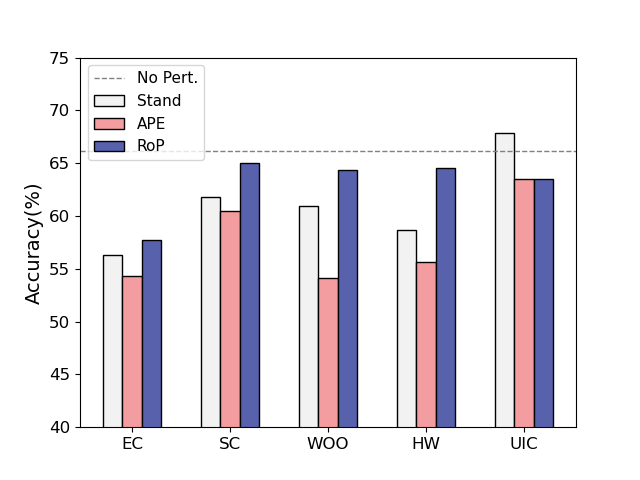}
    {(b) StrategyQA}
    \label{fig:subfig-StrategyQA}
  \end{minipage}
  \caption{The performance of RoP with different perturbation methods in the commonsense reasoning task.}
  \label{fig: cs}
\end{figure*}

\begin{figure*}
  \centering
  \begin{minipage}[b]{0.43\linewidth}
    \centering
    \includegraphics[width=\linewidth]{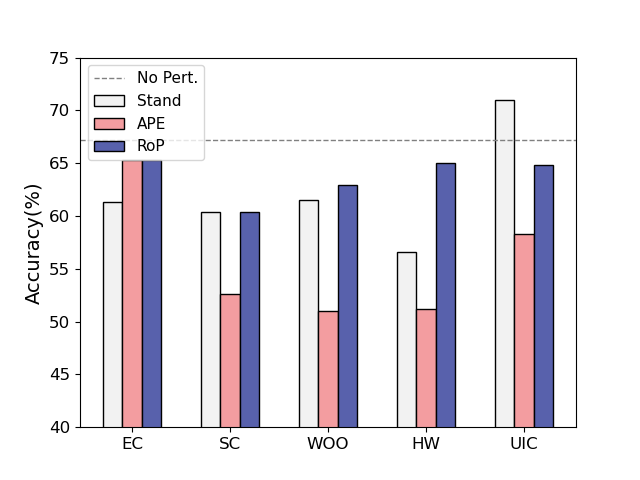}
    {(a) Date Understanding}
    \label{fig:subfig-date}
  \end{minipage}
  \hfill
  \begin{minipage}[b]{0.43\linewidth}
    \centering
    \includegraphics[width=\linewidth]{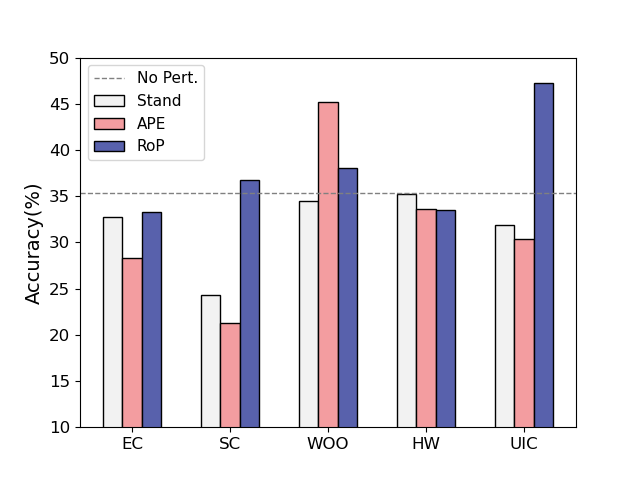}
    {(b) Tracking Shuffled Objects}
    \label{fig:subfig-track}
  \end{minipage}
  \caption{The performance of RoP with different perturbation methods in the logical reasoning task.}
  \label{fig: logical}
\end{figure*}

\subsection{Other Reasoning}
\subsubsection{Experiment Setup}
To further evaluate the generalization capacity of the RoP, we extended our experiments to include tasks in \textbf{Commonsense Reasoning} and \textbf{Logical Reasoning}. Specifically, we use two commonsense reasoning benchmarks and two logical reasoning tasks from BIG-bench~\cite{Bigbench}: 
\begin{itemize}
    \item \textbf{CSQA} (CommonsenseQA)~\cite{commonsenseqa} is a multiple‑choice benchmark dataset that presents questions with complex semantics, typically requiring reasoning over prior world knowledge.
    \item \textbf{StrategyQA}~\cite{StrategyQA} is a dataset that requires models to carry out implicit, multi‑step reasoning to arrive at an answer.
    \item \textbf{Date Understanding}~\cite{Bigbench} is a dataset that evaluates a model’s ability to infer specific dates from given contextual information.
    \item \textbf{Tracking Shuffled Objects}~\cite{Bigbench} is a dataset that evaluates a model’s ability to predict the final state of objects after being given an initial configuration and then subjected to a series of random reorderings. 
\end{itemize}

For these experiments, perturbations were applied to the test datasets by the protocol described in Section Method. Perturbed inputs, error correction instructions, and guidance prompts were generated using GPT-4o, while performance evaluations were conducted using GPT-3.5-Turbo.

\subsubsection{Result}
As illustrated in Fig.~\ref{fig: cs} and ~\ref{fig: logical}, RoP consistently outperforms both Stand and the APE method across nearly all datasets and perturbation types, demonstrating its robustness and domain-transferability. On CSQA dataset under EC perturbation, RoP achieves 71.3\% accuracy, outperforming APE (54.9\%) by +16.4\% and Stand (56.9\%) by +14.3\%. On Tracking Shuffled Objects under the UIC perturbation, RoP reaches 47.3\% accuracy, significantly surpassing APE (30.4\%) by +16.9\% and Stand (31.9\%) by +15.5\%.

These results confirm RoP’s robust performance in other tasks, affirming its utility as a general-purpose prompt for LLMs operating in adversarial contexts. An intriguing secondary finding emerged on StrategyQA and Date Understanding, where the inclusion of benign (irrelevant but non-adversarial) perturbations slightly improved model accuracy. This suggests that introducing peripheral or topic-adjacent noise may stimulate broader semantic activation in LLMs, thereby enhancing reasoning capacity under certain conditions.

\subsubsection{Using a Weaker Optimizer Model}
To further investigate the dependency of RoP on the capacity of the optimizer model, we substituted the optimizer with GPT-3.5-Turbo (denoted as GPT-3.5-Turbo Opt.). We then evaluated the performance of RoP on the AQUA, GSM8K, and CSQA datasets across all five perturbation types (EC, SC, WOO, HW, UIC). The performance was compared against the standard prompting baseline (Stand) and the original RoP configuration, which utilizes GPT-4o as the optimizer model (GPT-4o Opt.). The experimental results are presented in Table \ref{tab:supp_gpt35_r1}. The findings reveal that while replacing the optimizer with the weaker GPT-3.5-Turbo incurs a marginal performance degradation relative to GPT-4o, RoP consistently yields significant and stable robustness improvements across all perturbation scenarios. Furthermore, we observed that on the CSQA dataset under the EC and HW perturbation types, the performance achieved using GPT-3.5-Turbo Opt. slightly surpassed that of GPT-4o Opt.

These results indicate that the core framework of RoP retains strong effectiveness and cross-model generalizability, even when driven by lower-cost and smaller-scale models. Therefore, the efficacy of the proposed RoP strategy does not strictly depend on the utilization of highly capable large language models for instruction optimization.

\begin{table*}[!t]
    \centering
    \caption{Accuracy(\%) comparison of five perturbation methods with RoP when GPT-3.5-Turbo replaces GPT-4o as the optimization model on AQUA, GSM8K, and CSQA. No Dert.: indicates that the input questions have not been perturbed. $\Downarrow_{(*)}$: indicates that accuracy decreases compared to the Stand with No Dert. input.}
    \label{tab:supp_gpt35_r1}
    \begin{tabular}{cl|ccc|c}
        \toprule
        \textbf{Perturb}& \textbf{Method} & \textbf{AQUA} & \textbf{GSM8K} & \textbf{CSQA} & \textbf{Avg.}\\
        \midrule
        No Dert.& Stand & 74.02 & 75.66 & 77.56 & 75.75 \\
        \midrule
        \multirow{3}{*}{EC} & Stand & 57.87 & 69.07 & 56.92 & 61.29 $(\Downarrow_{14.46})$ \\
        & RoP (GPT-3.5-Turbo Opt.) & 62.20 & 68.61 & 73.52 & 68.11 $(\Downarrow_{7.64})$ \\
        & RoP (GPT-4o Opt.) & 62.99 & 72.86 & 71.25 & 69.03 $(\Downarrow_{6.72})$ \\
        \midrule
        \multirow{3}{*}{SC} & Stand & 52.36 & 68.01 & 70.76 & 63.71 $(\Downarrow_{12.04})$ \\
        & RoP (GPT-3.5-Turbo Opt.) & 60.24 & 72.40 & 72.38 & 68.34 $(\Downarrow_{7.41})$ \\
        & RoP (GPT-4o Opt.) & 70.08 & 75.36 & 74.94 & 73.46 $(\Downarrow_{2.29})$ \\
        \midrule
        \multirow{3}{*}{WOO} & Stand & 59.45 & 70.05 & 72.24 & 67.25 $(\Downarrow_{8.50})$ \\
        & RoP (GPT-3.5-Turbo Opt.) & 65.35 & 68.08 & 70.66 & 68.03 $(\Downarrow_{7.72})$ \\
        & RoP (GPT-4o Opt.) & 65.75 & 71.27 & 72.56 & 69.86 $(\Downarrow_{5.89})$ \\
        \midrule
        \multirow{3}{*}{HW} & Stand & 56.30 & 66.41 & 62.74 & 61.82 $(\Downarrow_{13.93})$ \\
        & RoP (GPT-3.5-Turbo Opt.) & 62.20 & 68.83 & 70.98 & 67.34 $(\Downarrow_{8.41})$ \\
        & RoP (GPT-4o Opt.) & 64.57 & 74.22 & 70.27 & 69.69 $(\Downarrow_{6.06})$ \\
        \midrule
        \multirow{3}{*}{UIC} & Stand & 57.09 & 51.86 & 74.77 & 61.24 $(\Downarrow_{14.51})$ \\
        & RoP (GPT-3.5-Turbo Opt.) & 60.24 & 56.10 & 69.53 & 61.96 $(\Downarrow_{13.79})$ \\
        & RoP (GPT-4o Opt.) & 62.20 & 61.71 & 70.84 & 64.92 $(\Downarrow_{10.83})$ \\
        \bottomrule
    \end{tabular}
\end{table*}

\subsubsection{Comparison with Basic Spelling Correction Tools}
In real-world scenarios, typographical errors represent a common source of input noise for Large Language Models (LLMs). Specifically, most perturbations—particularly Error Character (EC) and Similar Character (SC)—are closely related to conventional spelling errors. To rigorously address this and provide a direct baseline comparison, we conducted a new set of experiments on EC and SC perturbations. These experiments evaluate the performance of applying a basic spelling correction scheme prior to LLM inference and compare it directly with our proposed Robustness of Prompting (RoP) method.

For the spelling correction baseline, we employed the widely adopted Python library \texttt{pyspellchecker}. This is a pure-Python, dictionary-based tool directly built upon probabilistic spelling correction algorithms. The algorithm generates candidate words by considering all words within a Levenshtein edit distance of $ \le 2 $, and selects the highest-probability word based on a large English corpus. The correction process operates independently on each token, without relying on contextual understanding, task-specific knowledge, or the involvement of LLMs.

Our supplementary experiments were performed on the EC and SC perturbations across two arithmetic reasoning datasets (SingleEq and MultiArith) and one commonsense reasoning dataset (CSQA). For the spellchecker method, given a perturbed question $ \hat{x} $, the corrected input is obtained as follows:

\begin{equation}
\begin{aligned}
x_{\text{sc}} = \text{SpellChecker}(\hat{x})
\end{aligned}
\label{equ:spellcheck_correct}
\end{equation}

where each token is replaced by its closest dictionary match. The corrected input $ x_{\text{sc}} $ is then passed to the LLM to obtain its final answer $ \hat{y}_{\text{sc}} $:

\begin{equation}
\begin{aligned}
\hat{y}_{\text{sc}} = \arg\max_{y_i \in Y} L_\theta(y \mid x_{\text{sc}})
\end{aligned}
\label{equ:spellcheck_infer}
\end{equation}
In contrast, the RoP method was evaluated under its original experimental conditions, utilizing GPT-3.5-Turbo as the evaluator and GPT-4o as the instruction optimizer.

The experimental results are presented in Table~\ref{tab:spellchecker}. As demonstrated, RoP consistently outperforms the spelling correction baseline by a significant margin. In certain settings (e.g., SingleEq under EC perturbations), the dictionary-based method even results in an inference accuracy lower than that of the uncorrected Standard (Stand) baseline. This performance degradation occurs because general-purpose spelling checkers lack semantic and mathematical context: they may erroneously correct key domain-specific words to irrelevant dictionary entries or fail to properly handle special character substitutions, thereby introducing new errors that propagate into the reasoning phase. In comparison, RoP’s Error Correction stage performs contextual repair via auto-generated prompts, while the Guidance stage further steers the LLM's downstream inference process. These robustness capabilities remain fundamentally unattainable by purely lexical spelling correctors.

\begin{table}[!t]
\centering
\caption{Accuracy (\%) comparison of Stand, SpellChecker baseline, and RoP under EC and SC perturbations.}
\label{tab:spellchecker}
\begin{tabular}{llccc}
\hline
\textbf{Perturb} & \textbf{Method} & \textbf{SingleEq} & \textbf{MultiArith} & \textbf{CSQA} \\
\hline
\multirow{3}{*}{EC} 
& Stand         & 89.2\% & 87.0\% & 56.9\% \\
& SpellChecker  & 86.4\% & 92.1\% & 61.6\% \\
& RoP           & 95.3\% & 95.0\% & 71.3\% \\
\hline
\multirow{3}{*}{SC} 
& Stand         & 88.0\% & 85.3\% & 70.8\% \\
& SpellChecker  & 91.9\% & 91.2\% & 72.3\% \\
& RoP           & 95.1\% & 92.3\% & 74.9\% \\
\hline
\end{tabular}
\end{table}

\subsection{Inference Cost Analysis}
\subsubsection{Experiment Setup}
To enable a fair comparison of computational overhead, we adopt the average number of tokens per query---defined as the total tokens consumed by a specific perturbation-method combination divided by the total number of queries $N$---as the quantifiable metric for inference cost. Token usage is recorded directly from the OpenAI API. For each API call, the response provides \textbf{prompt\_tokens} (the number of tokens in the input prompt), \textbf{completion\_tokens} (the number of tokens in the generated output), and \textbf{total\_tokens} (their sum). To avoid ambiguity, we define the inference cost of a single API call as the \textbf{total\_tokens} throughout this paper.\\
We utilize the AQUA test set from our main experiments to ensure consistency with the robustness results reported in Table I and to preclude sampling bias. Each of the five perturbation types is independently evaluated on its corresponding perturbed question set. This design permits us to observe and report cost variations across different perturbation types, rather than relying on a single aggregate figure. The formal cost definitions for each method are as follows:\\
1. \textbf{Cost of RoP}: Let $\mathcal{D}_{\text{test}} = \{\hat{x}_i\}_{i=1}^N$ be the set containing $N = 254$ perturbed test questions. The total inference cost of RoP, denoted as $C_{\text{RoP}}$, is decomposed as:
\begin{equation}
C_{\text{RoP}} = C_{\text{in}_{\text{ec}}} + \sum_{i=1}^N C_{\text{corr}}(\hat{x}_i) + C_{\text{in}_{\text{opt}}} + \sum_{i=1}^N C_{\text{inf}}(x_{\text{ec},i}),
\label{equ:cost}
\end{equation}
where $C_{\text{in}_{\text{ec}}}$ and $C_{\text{in}_{\text{opt}}}$ are the one-time token costs for generating the error-correction instruction $in_{\text{ec}}$ and the guidance instruction $in_{\text{opt}}$, respectively. $C_{\text{corr}}(\hat{x}_i)$ represents the token cost of producing a single corrected input $x_{\text{ec},i}$ using the error-correction prompt $P = \langle in_{\text{ec}}, \hat{x}_i \rangle$. Finally, $C_{\text{inf}}(x_{\text{ec},i})$ is the token cost of generating the final answer via the inference prompt $P = \langle in_{\text{opt}}, \mathcal{D}_{\text{cor}}, x_{\text{ec},i} \rangle$.\\
2. \textbf{Costs of Baselines}: 
\begin{itemize}
    \item \textbf{Stand / CoT}: $C = \sum_{i=1}^N C_{\text{inf}}(\hat{x}_i)$ (inference overhead only; no optimization instruction cost).
    \item \textbf{APE}: $C = C_{\text{in}_{\text{opt}}} + \sum_{i=1}^N C_{\text{inf}}(\hat{x}_i)$.
    \item \textbf{PromptAgent}: $C_{\text{PromptAgent}} = C_{\text{opt}} + \sum_{i=1}^N C_{\text{inf}}(\hat{x}_i)$, where $C_{\text{opt}}$ is the token cost of the prompt optimization performed by PromptAgent.
\end{itemize}

\subsubsection{Result}
Table~\ref{tab:Costs} presents a summary of inference costs, juxtaposing the accuracy and average per-query cost of each method across the five perturbation types. To maintain conciseness, the complete token breakdown (\textbf{prompt\_tokens}, \textbf{completion\_tokens}, and \textbf{total\_tokens}) for every perturbation type is detailed in the Appendix.

The experimental results demonstrate that the per-query inference cost of RoP fluctuates between 1,377 and 1,760 tokens, depending on the perturbation type. This corresponds to approximately 1.74$\times$ to 3.00$\times$ the cost of the Stand baseline (579–848 tokens). For example, under the EC perturbation scenario, RoP incurs 1,493 tokens/query compared to Stand's 628 tokens/query, but it successfully raises the accuracy from 57.9\% to 63.0\% (+5.1 percentage points). Even in the highest-cost HW scenario, RoP achieves 64.6\% accuracy with 1,760 tokens, significantly outperforming Stand’s 56.3\%. Cost variation across different perturbation types is modest (a range of approximately 383 tokens for RoP), and its relative multiplier to the baseline remains stable. Furthermore, compared to APE (821–1,107 tokens/query) and CoT (1,057–1,105 tokens/query), RoP delivers superior robustness with a controllable overhead. PromptAgent, in contrast, incurs a cost exceeding 8,000 tokens/query, rendering it markedly less efficient.

These findings indicate that RoP attains exceptional robustness with only moderate inference overhead, well within the budget of practical LLM deployments. The analysis therefore confirms the strong practical value of RoP for real-world safety-critical applications that must jointly optimize accuracy and cost efficiency.

\begin{table}[!t]
\centering
\caption{Summary of Inference Costs on AQUA Dataset across All Perturbation Types.}
\label{tab:Costs}
\begin{tabular}{ll|rr}
\toprule
\textbf{Perturb} & \textbf{Method} & \textbf{Accuracy (\%)} & \textbf{Avg. Tokens/Query} \\
\midrule
\multirow{5}{*}{EC} 
& Stand & 57.9  & 628.0 \\
& CoT & 57.9  & 1104.2 \\
& APE & 57.9  & 1015.4 \\
& PromptAgent & 56.3  & 8874.5 \\
& \textbf{ROP (Our)} & 63.0  & 1493.2 \\
\midrule
\multirow{5}{*}{SC} 
& Stand & 52.4  & 602.7 \\
& CoT & 65.8  & 1105.3 \\
& APE & 52.8  & 916.4 \\
& PromptAgent & 58.7  & 18693.6 \\
& \textbf{ROP (Our)} & 70.1  & 1439.5 \\
\midrule
\multirow{5}{*}{WOO} 
& Stand & 59.5  & 579.4 \\
& CoT & 56.7  & 1056.8 \\
& APE & 60.6  & 821.1 \\
& PromptAgent & 57.1  & 15822.4 \\
& \textbf{ROP (Our)} & 65.8  & 1377.7 \\
\midrule
\multirow{5}{*}{HW} 
& Stand & 56.3  & 587.7 \\
& CoT & 63.4  & 1105.2 \\
& APE & 59.8  & 916.8 \\
& PromptAgent & 58.3  & 9099.4 \\
& \textbf{ROP (Our)} & 64.6  & 1759.9 \\
\midrule
\multirow{5}{*}{UIC} 
& Stand & 57.1  & 847.8 \\
& CoT & 50.0  & 1087.0 \\
& APE & 56.3  & 1106.7 \\
& PromptAgent & 48.4  & 9556.4 \\
& \textbf{ROP (Our)} & 62.2  & 1473.9 \\
\bottomrule
\end{tabular}
\end{table}

\section{Conclusion}
In this paper, we propose Robustness of Prompting (RoP), which is designed to enhance the robustness of LLMs. RoP consists of two stages: First is the \textit{Error Correction} stage, RoP generates adversarial examples by applying diverse perturbation methods to construct prompts that automatically correct the errors. Second is the \textit{Guidance} stage, RoP generates optimal guidance prompts based on the corrected input, steering the model’s inference process effectively. Our experiments demonstrate that RoP significantly improves LLMs' robustness against different types of adversarial perturbations, resulting in only minor performance degradation compared to unperturbed inputs.

\section{Acknowledgements}
This work is supported by the National Natural Science Foundation of China (No.62206004, No.62572002, No.62272001, No.624065095), the Natural Science Foundation of Anhui Province (No.2208085QF199, No.2508085MF159, No.2308085MF213), and the Hefei Key Technology R$\&$D ''Champion-Based Selection'' Project (No.2024SGJ010).

\section*{Appendix}
\subsection{Detailed Token Breakdown for Inference Cost Analysis}
This section details the comprehensive token consumption profiles (comprising prompt\_tokens, completion\_tokens, and total\_tokens) for RoP and all baseline models under the five distinct perturbation types in the AQUA dataset.

The detailed results are shown in Table~\ref{tab:token_breakdown_part1} and Table~\ref{tab:token_breakdown_part2}. Stand and CoT only include the query cost (Query), while APE and PromptAgent include the instruction generation cost (Inst.) and the query cost (Query). In contrast, the RoP method includes the instruction generation cost (Error Correction Instruction + Guidance Instruction), the error correction cost, and the query cost.

\begin{table}[htbp]
  \centering
  \caption{Detailed Token Consumption Statistics for Different Methods Under Various Perturbation Types (AQuA Dataset) (Part I)}
  \label{tab:token_breakdown_part1}
  \resizebox{\linewidth}{!}{
  \begin{tabular}{lllrcrr}
    \toprule
    \textbf{Method} & \textbf{Pert.} & \textbf{Token Type} & \textbf{Inst.} & \textbf{Correction} & \textbf{Query} & \textbf{Total} \\
    \midrule
    
    \multirow{15}{*}{Stand} 
    & \multirow{3}{*}{EC} 
    & Prompt & -- & -- & 34,971 & 34,971 \\
    & & Completion & -- & -- & 124,531 & 124,531 \\
    & & Total & -- & -- & 159,502 & 159,502 \\
    \cmidrule{2-7}
    & \multirow{3}{*}{SC} 
    & Prompt & -- & -- & 32,089 & 32,089 \\
    & & Completion & -- & -- & 120,993 & 120,993 \\
    & & Total & -- & -- & 153,082 & 153,082 \\
    \cmidrule{2-7}
    & \multirow{3}{*}{WOO} 
    & Prompt & -- & -- & 22,431 & 22,431 \\
    & & Completion & -- & -- & 124,734 & 124,734 \\
    & & Total & -- & -- & 147,165 & 147,165 \\
    \cmidrule{2-7}
    & \multirow{3}{*}{HW} 
    & Prompt & -- & -- & 23,245 & 23,245 \\
    & & Completion & -- & -- & 126,027 & 126,027 \\
    & & Total & -- & -- & 149,272 & 149,272 \\
    \cmidrule{2-7}
    & \multirow{3}{*}{UIC} 
    & Prompt & -- & -- & 30,782 & 30,782 \\
    & & Completion & -- & -- & 184,570 & 184,570 \\
    & & Total & -- & -- & 215,352 & 215,352 \\
    \midrule

    \multirow{15}{*}{COT} 
    & \multirow{3}{*}{EC} 
    & Prompt & -- & -- & 166,972 & 166,972 \\
    & & Completion & -- & -- & 113,484 & 113,484 \\
    & & Total & -- & -- & 280,456 & 280,456 \\
    \cmidrule{2-7}
    & \multirow{3}{*}{SC} 
    & Prompt & -- & -- & 163,678 & 163,678 \\
    & & Completion & -- & -- & 117,065 & 117,065 \\
    & & Total & -- & -- & 280,743 & 280,743 \\
    \cmidrule{2-7}
    & \multirow{3}{*}{WOO} 
    & Prompt & -- & -- & 158,279 & 158,279 \\
    & & Completion & -- & -- & 110,156 & 110,156 \\
    & & Total & -- & -- & 268,435 & 268,435 \\
    \cmidrule{2-7}
    & \multirow{3}{*}{HW} 
    & Prompt & -- & -- & 172,280 & 172,280 \\
    & & Completion & -- & -- & 108,437 & 108,437 \\
    & & Total & -- & -- & 280,717 & 280,717 \\
    \cmidrule{2-7}
    & \multirow{3}{*}{UIC} 
    & Prompt & -- & -- & 158,864 & 158,864 \\
    & & Completion & -- & -- & 117,243 & 117,243 \\
    & & Total & -- & -- & 276,107 & 276,107 \\
    \midrule

    \multirow{15}{*}{APE} 
    & \multirow{3}{*}{EC} 
    & Prompt & 67,068 & -- & 42,595 & 109,663 \\
    & & Completion & 13,942 & -- & 134,317 & 148,259 \\
    & & Total & 81,010 & -- & 176,912 & 257,922 \\
    \cmidrule{2-7}
    & \multirow{3}{*}{SC} 
    & Prompt & 63,069 & -- & 36,935 & 100,004 \\
    & & Completion & 14,873 & -- & 117,898 & 132,771 \\
    & & Total & 77,942 & -- & 154,833 & 232,775 \\
    \cmidrule{2-7}
    & \multirow{3}{*}{WOO} 
    & Prompt & 51,304 & -- & 26,746 & 78,050 \\
    & & Completion & 14,447 & -- & 116,062 & 130,509 \\
    & & Total & 65,751 & -- & 142,808 & 208,559 \\
    \cmidrule{2-7}
    & \multirow{3}{*}{HW} 
    & Prompt & 52,052 & -- & 28,057 & 80,109 \\
    & & Completion & 14,189 & -- & 138,576 & 152,765 \\
    & & Total & 66,241 & -- & 166,633 & 232,874 \\
    \cmidrule{2-7}
    & \multirow{3}{*}{UIC} 
    & Prompt & 63,405 & -- & 36,618 & 100,023 \\
    & & Completion & 14,959 & -- & 166,126 & 181,085 \\
    & & Total & 78,364 & -- & 202,744 & 281,108 \\
    \bottomrule
  \end{tabular}}
\end{table}

\begin{table}[htbp]
  \centering
  \caption{Detailed Token Consumption Statistics for Different Methods Under Various Perturbation Types (AQuA Dataset) (Part II)}
  \label{tab:token_breakdown_part2}
  \setlength{\tabcolsep}{3pt}
  \resizebox{\linewidth}{!}{
  \begin{tabular}{lllrcrr}
    \toprule
    \textbf{Method} & \textbf{Pert.} & \textbf{Token Type} & \textbf{Inst.} & \textbf{Correction} & \textbf{Query} & \textbf{Total} \\
    \midrule

    \multirow{15}{*}{PromptAgent} 
    & \multirow{3}{*}{EC} 
    & Prompt & 796,507 & -- & 77,644 & 874,151 \\
    & & Completion & 1,209,297 & -- & 170,664 & 1,379,961 \\
    & & Total & 2,005,804 & -- & 248,308 & 2,254,112 \\
    \cmidrule{2-7}
    & \multirow{3}{*}{SC} 
    & Prompt & 1,504,661 & -- & 154,438 & 1,659,099 \\
    & & Completion & 2,749,744 & -- & 339,340 & 3,089,084 \\
    & & Total & 4,254,405 & -- & 493,778 & 4,748,183 \\
    \cmidrule{2-7}
    & \multirow{3}{*}{WOO} 
    & Prompt & 1,318,260 & -- & 187,274 & 1,505,534 \\
    & & Completion & 1,971,358 & -- & 542,007 & 2,513,365 \\
    & & Total & 3,289,618 & -- & 729,281 & 4,018,899 \\
    \cmidrule{2-7}
    & \multirow{3}{*}{HW} 
    & Prompt & 791,906 & -- & 53,534 & 845,440 \\
    & & Completion & 1,310,168 & -- & 155,636 & 1,465,804 \\
    & & Total & 2,102,074 & -- & 209,170 & 2,311,244 \\
    \cmidrule{2-7}
    & \multirow{3}{*}{UIC} 
    & Prompt & 791,867 & -- & 72,268 & 864,135 \\
    & & Completion & 1,291,909 & -- & 271,280 & 1,563,189 \\
    & & Total & 2,083,776 & -- & 343,548 & 2,427,324 \\
    \midrule

    \multirow{15}{*}{ROP} 
    & \multirow{3}{*}{EC} 
    & Prompt & 57,378 + 90,493 & 41,322 & 26,831 & 216,024 \\
    & & Completion & 15,232 + 14,319 & 21,489 & 112,203 & 163,243 \\
    & & Total & 72,610 + 104,812 & 62,811 & 139,034 & 379,267 \\
    \cmidrule{2-7}
    & \multirow{3}{*}{SC} 
    & Prompt & 57,090 + 87,792 & 39,221 & 28,742 & 212,845 \\
    & & Completion & 14,289 + 13,451 & 27,778 & 97,272 & 152,790 \\
    & & Total & 71,379 + 101,243 & 66,999 & 126,014 & 365,635 \\
    \cmidrule{2-7}
    & \multirow{3}{*}{WOO} 
    & Prompt & 56,785 + 77,695 & 28,270 & 28,548 & 191,298 \\
    & & Completion & 15,284 + 14,110 & 22,176 & 107,063 & 158,633 \\
    & & Total & 72,069 + 91,805 & 50,446 & 135,611 & 349,931 \\
    \cmidrule{2-7}
    & \multirow{3}{*}{HW} 
    & Prompt & 57,142 + 73,909 & 27,803 & 27,565 & 186,419 \\
    & & Completion & 15,178 + 12,318 & 21,655 & 211,454 & 260,605 \\
    & & Total & 72,320 + 86,227 & 49,458 & 239,019 & 447,024 \\
    \cmidrule{2-7}
    & \multirow{3}{*}{UIC} 
    & Prompt & 50,049 + 91,046 & 39,500 & 28,722 & 209,317 \\
    & & Completion & 14,665 + 13,845 & 25,244 & 111,293 & 165,047 \\
    & & Total & 64,714 + 104,891 & 64,744 & 140,015 & 374,364 \\
    \bottomrule
  \end{tabular}}
\end{table}

\subsection{Detailed Prompt Structures (AQUA)}
\textbf{1. EC}

1) RoP Error Correction Prompt:

\begin{tcolorbox}[colback=white, colframe=black, boxrule=0.5pt]
Correct the spelling errors and improve the readability of the text while preserving the original meaning and formatting, including the answer choices.
\end{tcolorbox}

2) RoP Guidance Prompt:

\begin{tcolorbox}[colback=white, colframe=black, boxrule=0.5pt]
Solve the given mathematical or logical problem and choose the correct answer from the provided multiple-choice options.
\end{tcolorbox}

3) APE Prompt:

\begin{tcolorbox}[colback=white, colframe=black, boxrule=0.5pt]
Solve problems with scrambled or misspelled text and select the correct answer from the given choices.
\end{tcolorbox}

4) promptagent Prompt:

\begin{tcolorbox}[colback=white, colframe=black, boxrule=0.5pt]
Please solve the following mathematical word problems. For each problem, ensure your response includes all necessary calculations, clearly presented to logically lead to the final answer. After completing your calculations, provide the final answer formatted as $<$answer$>$your\_answer\_here$<$/answer$>$. If the answer is an integer, confirm it is the closest valid option from the answer choices provided. Make sure to cross-check your logic and results for consistency, and explicitly state the answer choice label (A, B, C, D, E) that corresponds to your final answer before $<$answer$>$. If your final answer does not match any of the provided choices, clearly indicate the nearest valid option based on your calculations.
\end{tcolorbox}

\textbf{2. SC}

1) RoP Error Correction Prompt:

\begin{tcolorbox}[colback=white, colframe=black, boxrule=0.5pt]
Normalize the text by correcting any misspellings, converting special characters to standard ones, and ensuring proper formatting of mathematical expressions and answer choices.
\end{tcolorbox}

2) RoP Guidance Prompt:

\begin{tcolorbox}[colback=white, colframe=black, boxrule=0.5pt]
Solve mathematical and logical problems and choose the correct answer from the provided options.
\end{tcolorbox}

3) APE Prompt:

\begin{tcolorbox}[colback=white, colframe=black, boxrule=0.5pt]
Create math word problems and provide multiple choice answers, ensuring the final correct answer matches the given output.
\end{tcolorbox}

4) promptagent Prompt:

\begin{tcolorbox}[breakable, colback=white, colframe=black, boxrule=0.5pt]
Please solve the following mathematical word problems with precision and clarity. Your response should include detailed step-by-step calculations, logical reasoning, and a clear identification of the correct answer from the provided options. 

1. If your calculated answer does not exactly match any of the options, select the closest one and clearly justify your choice, explaining how you arrived at that conclusion.

2. Be particularly vigilant about ensuring that all numerical outputs are presented as integers where applicable, especially when relevant to whole numbers.

3. After each significant calculation, explicitly state how it relates to the available answer choices, demonstrating a connection between your calculations and the options.

4. At the end of your response, format your final answer by enclosing it in the tags $<$answer$>$ and $<$/answer$>$, ensuring that you also indicate which answer choice (A, B, C, D, E) your final answer corresponds to. 

5. Provide a concise closing summary that reiterates your findings and clarifies any mathematical concepts or relationships that are relevant to the problems addressed.

By following these guidelines, you can produce accurate, well-justified solutions throughout the process.
\end{tcolorbox}

\textbf{3. WOO}

1) RoP Error Correction Prompt:

\begin{tcolorbox}[colback=white, colframe=black, boxrule=0.5pt]
Rewrite the given math word problems into clearer and more grammatically correct sentences while retaining the original information and answer choices.
\end{tcolorbox}

2) RoP Guidance Prompt:

\begin{tcolorbox}[colback=white, colframe=black, boxrule=0.5pt]
Select the correct answer choice for each problem statement by solving or calculating the problem provided in the input.
\end{tcolorbox}

3) APE Prompt:

\begin{tcolorbox}[colback=white, colframe=black, boxrule=0.5pt]
Solve the given word problems and select the correct answer choice.
\end{tcolorbox}

4) promptagent Prompt:

\begin{tcolorbox}[breakable,colback=white, colframe=black, boxrule=0.5pt]
Please solve the following mathematical word problems. For each problem, follow these guidelines for clarity and accuracy:

1. Provide a step-by-step explanation of your reasoning and any calculations involved. Make sure to define any variables and summarize the problem appropriately before proceeding.
  
2. If the problem has specific components (e.g., "only summer vacation" or "only last year’s data"), ensure you clearly distinguish these during your explanation.

3. When presenting your final answer, enclose it within the tags $<$answer$>$ and $<$/answer$>$ and include the corresponding label from the answer choices provided (A, B, C, D, etc.) immediately before the answer tag.

4. If your calculations involve rounding or estimating, clearly indicate this in your response.

5. Be concise and maintain clarity throughout your explanation to avoid misunderstandings with the problem statement.

Thank you!
\end{tcolorbox}

\textbf{4. HW}

1) RoP Error Correction Prompt:

\begin{tcolorbox}[colback=white, colframe=black, boxrule=0.5pt]
Sorrect spelling and grammatical errors in mathematical problem statements while preserving the structure and multiple-choice format.
\end{tcolorbox}

2) RoP Guidance Prompt:

\begin{tcolorbox}[colback=white, colframe=black, boxrule=0.5pt]
Solve multiple-choice mathematical problem statements and select the correct answer choice for each problem.
\end{tcolorbox}

3) APE Prompt:

\begin{tcolorbox}[colback=white, colframe=black, boxrule=0.5pt]
Solve word problems with deliberate spelling errors, choosing the correct answer from multiple-choice options.
\end{tcolorbox}

4) promptagent Prompt:

\begin{tcolorbox}[colback=white, colframe=black, boxrule=0.5pt]
Please solve the following mathematical word problem step-by-step. Clearly indicate each calculation and keep the variable names as they are without modification. Once you have derived the solution, select the largest integer from the given options that satisfies the condition of the problem. Label your selection as follows: Final answer: $<$answer$>$(A), (B), (C), (D), or (E)$<$/answer$>$. If none of the answer choices is valid, indicate: "None of the options provided contain the correct answer." Ensure your final answer is formatted correctly and included in the specified brackets.
\end{tcolorbox}

\textbf{5. UIC}

1) RoP Error Correction Prompt:

\begin{tcolorbox}[colback=white, colframe=black, boxrule=0.5pt]

Simplify the problem statements by removing any extraneous details that do not impact the primary question or required calculations, while maintaining the essential information needed to answer the question correctly.
\end{tcolorbox}

2) RoP Guidance Prompt:

\begin{tcolorbox}[colback=white, colframe=black, boxrule=0.5pt]
Simplify the problem statement and identify the correct answer from the given choices.
\end{tcolorbox}

3) APE Prompt:

\begin{tcolorbox}[colback=white, colframe=black, boxrule=0.5pt]

Solve the given mathematical and logical problems and select the correct answer choice from the options provided.
\end{tcolorbox}

4) promptagent Prompt:
\begin{tcolorbox}[colback=white, colframe=black, boxrule=0.5pt]
Please solve the following mathematical word problems by clearly identifying and analyzing the relationships involved (such as ratios, percentages, and proportions). Provide detailed calculations for your answers, and ensure that the final answer is presented in a format that corresponds to the provided answer choices. Include each step of your reasoning to clarify how you arrived at the solution, and double-check that your final answer aligns with the options given.
\end{tcolorbox}

\bibliographystyle{ieeetr}

\bibliography{TAI_template}

\end{document}